\documentclass{article}

\usepackage{microtype}
\usepackage{graphicx}
\usepackage{booktabs} 

\usepackage{hyperref}

\usepackage[final,nonatbib]{neurips_2018}

\usepackage[utf8]{inputenc} 
\usepackage[T1]{fontenc}    
\usepackage{hyperref}       
\usepackage{url}            
\usepackage{booktabs}       
\usepackage{amsfonts}       
\usepackage{nicefrac}       
\usepackage{microtype}      

\usepackage[numbers]{natbib}

\usepackage{subfiles}
\usepackage{amsmath}
\usepackage{amsthm}
\usepackage{amssymb}
\usepackage{algorithm}
\usepackage{algpseudocode}
\usepackage{mathdots}
\usepackage{times}
\usepackage{paralist}
\usepackage{anyfontsize}
\usepackage[svgnames]{xcolor}
\usepackage{relsize}
\hypersetup{citecolor=blue,colorlinks=true,breaklinks=true,urlcolor=blue}
\usepackage{tikz}
\usepackage{stfloats}
\usepackage{xr}
\usepackage{mathtools}
\usepackage{grffile}
\usetikzlibrary{arrows, chains, fit, shapes, decorations.pathreplacing, decorations.pathmorphing, positioning, calc, backgrounds, hobby}

\usepackage{caption}
\usepackage{subcaption}
\usepackage{wrapfig}
\usepackage{float}
\usepackage{enumitem}
\newfloat{algorithm}{t}{lop}

\tikzset{square arrow/.style={to path={-- ++(0,.15) -| (\tikztotarget)}}}
\tikzstyle{bayes_net}=[->, >=stealth', shorten >=1pt, auto, node distance=2cm, thick]
\tikzstyle{markov_random_field}=[>=stealth', auto, node distance = 2cm, thick]
\tikzstyle{main_node}=[circle, rounded corners, draw, minimum size = 0.85cm]
\tikzstyle{dummy_node}=[circle, rounded corners, minimum size = 0.85cm]
\tikzstyle{box_node}=[draw, minimum size = 0.7cm]

\theoremstyle{definition}
\newtheorem*{definition}{Definition}

\DeclarePairedDelimiterX{\infdivx}[2]{(}{)}{%
  #1\;\delimsize\|\;#2%
}
\newcommand{\kl}{D\infdivx}
\newcommand{\supp}{\text{supp}}

\renewcommand{\algorithmicrequire}{\textbf{Input:}}
\renewcommand{\algorithmicensure}{\textbf{Output:}}

\newcommand{\ie}{\emph{i.e.}}
\newcommand{\etc}{\emph{etc}}
\newcommand{\eg}{\emph{e.g.}}
\newcommand{\hide}[1]{}
\newcommand{\todo}[1]{ {\color{red} \textbf{TODO:} {#1}} }

\newcommand{\N}{\mathcal{N}}

\graphicspath{{./}{./plots/}}

\title{Meta-Learning MCMC Proposals}

\author{
  \makebox[0.33\textwidth][c]{Tongzhou Wang\thanks{Work done while the author was at the University of California, Berkeley}} \\
  \makebox[0.33\textwidth][c]{Facebook AI Research} \\
  \makebox[0.33\textwidth][c]{\texttt{tongzhou.wang.1994@gmail.com}} \\
  \And
  \makebox[0.33\textwidth][c]{Yi Wu} \\
  \makebox[0.33\textwidth][c]{University of California, Berkeley}\\
  \makebox[0.33\textwidth][c]{\texttt{jxwuyi@gmail.com}} \\
  \And
  \makebox[0.33\textwidth][c]{David A. Moore\thanks{Work done while the author was at the University of California, Berkeley}} \\
  \makebox[0.33\textwidth][c]{Google} \\
  \makebox[0.33\textwidth][c]{\texttt{davmre@gmail.com}} \\
  \And
  \makebox[0.33\textwidth][c]{Stuart J. Russell} \\
  \makebox[0.33\textwidth][c]{University of California, Berkeley}\\
  \makebox[0.33\textwidth][c]{\texttt{russell@cs.berkeley.edu}} \
}

\begin{document}
\maketitle

\begin{abstract}

\hide{
 Effective implementations of sampling-based probabilistic inference often require manually constructed, model-specific proposals. 
 We propose a meta-learning approach to automate proposal construction by training neural networks to provide fast approximations to block Gibbs conditionals. The learned neural proposals generalize to occurrences of common structural motifs across different models, allowing for the construction of a library of learned inference primitives that can accelerate inference on unseen models with no model-specific training required. We explore several applications including open-universe Gaussian mixture models, in which our learned proposals outperform a hand-tuned sampler, and a real-world named entity recognition task, in which our sampler yields higher final F1 scores than classical single-site Gibbs sampling.
 }

Effective implementations of sampling-based probabilistic inference often require manually constructed, model-specific proposals. Inspired by recent progresses in meta-learning for training learning agents that can generalize to unseen environments, we propose a meta-learning approach to building effective and generalizable MCMC proposals. 
We parametrize the proposal as a neural network to provide fast approximations to block Gibbs conditionals. 
The learned neural proposals generalize to occurrences of common structural motifs across different models, allowing for the construction of a library of learned inference primitives that can accelerate inference on unseen models with no model-specific training required. We explore several applications including open-universe Gaussian mixture models, in which our learned proposals outperform a hand-tuned sampler, and a real-world named entity recognition task, in which our sampler yields higher final F1 scores than classical single-site Gibbs sampling.

\hide{sampler's ability to escape local modes}

\end{abstract}

\section{Introduction}\label{sec:intro}
\input{10intro}

\section{Related Work}\label{sec:related}
There has been great interest in using learned, feedforward inference networks to generate approximate posteriors. Variational autoencoders (VAE) train an inference network jointly with the parameters of the forward model to maximize a variational lower bound \citep{kingma2013auto,burda2015importance,gu2015neural}. However, the use of a parametric variational distribution means they typically have limited capacity to represent complex, potentially multimodal posteriors, such as those incorporating discrete variables or structural uncertainty.

A related line of work has developed data-driven proposals for importance samplers \citep{paige2016inference,le2017inference,ritchie2016guided}, training an inference network from prior samples which is then used as a proposal given observed evidence. In particular, \citet{le2017inference} generalize the framework to probabilistic programming, and is able to automatically generate and train a neural proposal network given an arbitrary model described in a probabilistic program. Our approach differs in that we focus on MCMC inference, allowing modular proposals for subsets of model variables that may depend on latent quantities, and exploit recurring structural motifs to generalize to new models with no additional training.

Several approaches have been proposed for adaptive block sampling, in which sets of variables exhibiting strong correlations are identified dynamically during inference, so that costly joint sampling is used only for blocks where it is likely to be beneficial \cite{venugopal2013dynamic,turek2016automated}. This is largely complementary to our current approach, which assumes the set of blocks (structural motifs) is given and attempts to learn fast approximate proposals.

Perhaps most related to our approach is recent work that trains model-specific MCMC proposals with machine learning techniques. In \cite{song2017nice}, adversarial training directly optimizes the similarity between posterior values and proposed values from a symmetric MCMC proposal. 
Stochastic inverses of graphical models\cite{stuhlmuller2013learning} train density estimators to speed up inference. However, both approaches have limitations on applicable models and require model-specific training using global information (samples containing all variables). Our approach is simpler and more scalable, requiring only local information and generating local proposals that can be reused both within and across different models. 

At a high level, our approach of learning an approximate local update scheme can be seen as related to approximate message passing \citep{ross2011learning, heess2013learning} and learning to optimize continuous objectives \citep{andrychowicz2016learning,li2017learning}.


\section{Meta-Learning MCMC Proposals}\label{sec:neural}

We propose a meta-learning approach, using a neural network to approximate the Gibbs proposal for a recurring structural motif in graphical models, and to speed up inference on unseen models without extra tuning. Crucially our proposals do {\em not} fix the model parameters, which are instead provided as network input. After training with random model parametrizations, the same trained proposal can be reused to perform inference on novel models with parametrizations not previously observed.

Our inference networks are parametrized as mixture density networks \citep{bishop1994mixture}, and trained to minimize the Kullback{-}Leibler (KL) divergence between the true posterior conditional and the proposal by sampling instantiations of the motif. The proposals are then accepted or rejected following the Metropolis-Hastings (MH) rule \citep{andrieu2003introduction}, so we maintain the correct stationary distribution even though the proposals are approximate. The following sections describe our work in greater depth.

\hide{Sec.~\ref{sec:back} outlines the basic idea and background of our approach. Sec.~\ref{sec:motif} discusses the idea of applying neural block proposals on structural motifs. Sec.~\ref{sec:param} and Sec.~\ref{sec:neural-train} show how we use and train neural networks to parametrize proposals. Finally, Sec.~\ref{sec:mcmc} describes the usual settings and steps to apply our method in practical scenarios.}

\subsection{Background}\label{sec:back}

Although our approach applies to arbitrary probabilistic programs, for simplicity we focus on models represented as factor graphs. A model consists of a set of variables $V$ as the nodes of a graph $G = (V, E)$, along with a set of factors specifying a joint probability distribution $p_\Psi(V)$ described by parameters $\Psi$. In particular, this paper focuses primarily on directed models, in which the factors $\Psi$ specify the conditional probability distributions of each variable given its parents. 
In undirected models, such as the Conditional Random Fields (CRFs) in Sec.~\ref{sec:ner}, the factors are arbitrary functions associated with cliques in the graph \citep{koller2009probabilistic}.

Given a set of observed evidence variables, inference attempts to sample from the conditional distribution on the remaining variables. %
\hide{A standard approach is Gibbs sampling \citep{spiegelhalter1996bugs,andrieu2003introduction}, in which each variable $v_i$ is successively resampled from its conditional distribution $p(v_i | V_{\neg i})$ given all other variables $V_{\neg i}$ in the graph. In most cases this conditional depends only on a subset of $V_{\neg i}$, known as the {\em Markov blanket}, $\text{MB}(v_i) \subseteq V_{\neg i}$. Each Gibbs update can be viewed as a MH proposal that is accepted by construction, thus inheriting the MH guarantee that the limiting distribution of the sampling process is the desired posterior conditional distribution \citep{andrieu2003introduction}.}%
In order to construct good MCMC proposals that generalize well across a variety of inference tasks, we take the advantage of recurring {\em structural motifs} in graphical models, such as grids, rings, and chains \citep{kemp2008discovery}.

In this work, our goal is to train a neural network as an efficient expert proposal for a structural motif, with its inputs containing the local parameters, so that the trained proposal can be applied to different models. Within a motif, the variables are divided into a proposed set of variables that will be resampled, and a conditioning set corresponding to an approximate Markov blanket. The proposal network essentially maps the values of conditional variables and local parameters to a distribution over the proposed variables.

\subsection{MCMC Proposals on Structural Motifs in Graphical Models}\label{sec:motif}

\begin{figure}
    \centering
    \begin{subfigure}[t]{0.48\linewidth}
        \centering
        \scalebox{0.87}{
        \begin{tikzpicture}[bayes_net, node distance = 0.65cm]
            \node[main_node, minimum size = 0.35cm] (3) {};
            \node[main_node, minimum size = 0.35cm] (2) [left = of 3] {};
            \node[main_node, minimum size = 0.35cm, fill=black!20] (1) [left = of 2] {};
            \node[main_node, minimum size = 0.35cm, fill=black!20] (0) [left = of 1] {};
            \node[main_node, minimum size = 0.35cm, fill=black!20] (4) [right = of 3] {};
            \node[main_node, minimum size = 0.35cm, fill=black!20] (5) [right = of 4] {};
            \node[main_node, minimum size = 0.35cm] (6) [right = of 5] {};

            \path[every node/.style = {font = \sffamily\small}]
            (0) edge [right] node [left] {} (1)
            (0) edge [bend left] node [left] {} (2)
            (1) edge [right] node [left] {} (2)
            (1) edge [bend right] node [left] {} (3)
            (2) edge [right] node [left] {} (3)
            (2) edge [bend left] node [left] {} (4)
            (3) edge [right] node [left] {} (4)
            (3) edge [bend right] node [left] {} (5)
            (4) edge [right] node [left] {} (5)
            (4) edge [bend left] node [left] {} (6)
            (5) edge [right] node [left] {} (6)
            ;

            \node [yshift = 0.25ex, xshift = 0ex] (0t) at (0.north) {};
            \node [yshift = -0.25ex, xshift = 0ex] (0b) at (0.south) {};
            \node (motif1) [draw = black, fit = (0t) (0b) (0) (1) (2) (3) (4) (5), draw opacity = 0.5, dashed, rounded corners = 0.35cm, inner sep = 0.2cm, rectangle] {};

            \node [yshift = 0.25ex, xshift = 0ex] (6t) at (6.north) {};
            \node [yshift = -0.25ex, xshift = 0ex] (6b) at (6.south) {};
            \node (motif2) [fit = (6t) (6b) (1) (2) (3) (4) (5) (6), draw opacity = 0.0, dashed, rounded corners = 0.35cm, inner sep = 0.2cm, rectangle] {};
        \end{tikzpicture}
        }
        \caption{One instantiation.}
    \end{subfigure}\hfill%
    \begin{subfigure}[t]{0.48\linewidth}
        \centering
        \scalebox{0.87}{
        \begin{tikzpicture}[bayes_net, node distance = 0.65cm]
            \node[main_node, minimum size = 0.35cm] (3) {};
            \node[main_node, minimum size = 0.35cm, fill=black!20] (2) [left = of 3] {};
            \node[main_node, minimum size = 0.35cm, fill=black!20] (1) [left = of 2] {};
            \node[main_node, minimum size = 0.35cm] (0) [left = of 1] {};
            \node[main_node, minimum size = 0.35cm] (4) [right = of 3] {};
            \node[main_node, minimum size = 0.35cm, fill=black!20] (5) [right = of 4] {};
            \node[main_node, minimum size = 0.35cm, fill=black!20] (6) [right = of 5] {};

            \path[every node/.style = {font = \sffamily\small}]
            (0) edge [right] node [left] {} (1)
            (0) edge [bend left] node [left] {} (2)
            (1) edge [right] node [left] {} (2)
            (1) edge [bend right] node [left] {} (3)
            (2) edge [right] node [left] {} (3)
            (2) edge [bend left] node [left] {} (4)
            (3) edge [right] node [left] {} (4)
            (3) edge [bend right] node [left] {} (5)
            (4) edge [right] node [left] {} (5)
            (4) edge [bend left] node [left] {} (6)
            (5) edge [right] node [left] {} (6)
            ;

            \node [yshift = 0.25ex, xshift = 0ex] (0t) at (0.north) {};
            \node [yshift = -0.25ex, xshift = 0ex] (0b) at (0.south) {};
            \node (motif1) [fit = (0t) (0b) (0) (1) (2) (3) (4) (5), draw opacity = 0.0, dashed, rounded corners = 0.35cm, inner sep = 0.2cm, rectangle] {};

            \node [yshift = 0.25ex, xshift = 0ex] (6t) at (6.north) {};
            \node [yshift = -0.25ex, xshift = 0ex] (6b) at (6.south) {};
            \node (motif2) [draw = black, fit = (6t) (6b) (1) (2) (3) (4) (5) (6), draw opacity = 0.5, dashed, rounded corners = 0.35cm, inner sep = 0.2cm, rectangle] {};
        \end{tikzpicture}
        }
        \caption{Another instantiation.}
    \end{subfigure}
    \caption{Two instantiations of a structural motif in a directed chain of length seven. The motif consists of two consecutive variables and their Markov blanket of four neighboring variables. Each instantiation is separated into block proposed variables $B_i$ (white) and conditioning variables $C_i$ (shaded).}
    \label{fig:motif-example}
\end{figure}
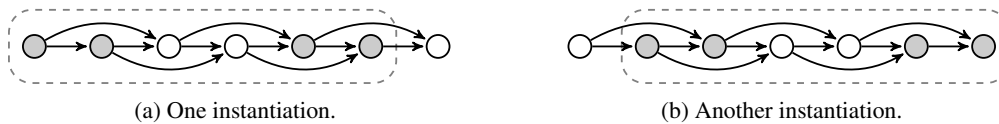

We associate each learned proposal with a {\em structural motif} that determines the shape of the network inputs and outputs. In general, structural motifs can be arbitrary subgraphs, but we are more interested in motifs that represent interesting conditional structure between two sets of variables, the block proposed variables $B$ and the conditioning variables $C$. A given motif can have multiple instantiations with a model, or even across models. As a concrete example, Fig.~\ref{fig:motif-example} shows two instantiations of a structural motif of six consecutive variables in a chain model. In each instantiation, we want to approximate the conditional distribution of two middle variables given neighboring four.%

%
\begin{definition} \label{def:sm}%
    A structural motif $(B, C)$ (or motif in short) is an (abstract) graph with nodes partitioned into two sets, $B$ and $C$, and a parametrized joint distribution $p(B, C)$ whose factorization is consistent with the graph structure. This specifies the functional form of the conditional $p(B|C)$, but not the specific parameters.
\end{definition}\vspace{-4pt}
A motif usually have many instantiations across many different graphical models.

\begin{definition} \label{def:sm_instantiation}
    For a graphical model $(G = (V, E), \Psi)$, an {\em instantiation} $(B_i, C_i, \Psi_i)$ of a motif $(B, C)$ includes%
    \begin{enumerate}[leftmargin=15pt,topsep=-3.5pt,itemsep=0.2ex,partopsep=5pt,parsep=0ex]
        \item a subset of the model variables $(B_i, C_i) \subseteq V$ such that the induced subgraph on $(B_i, C_i)$ is isomorphic to the motif $(B, C)$ with the partition preserved by the isomorphism (so nodes in $B$ are mapped to $B_i$, and $C$ to $C_i$), and
        \item a subset of model parameters $\Psi_i \subseteq \Psi$ required to specify the conditional distribution $p_{\Psi_i}(B | C)$.
    \end{enumerate}
\end{definition}\vspace{-4pt}

We would typically define a structural motif by first picking out a block of variables $B$ to jointly sample, and then selecting a conditioning set $C$. Intuitively, the natural choice for a conditioning set is the Markov blanket, $C = \text{MB}(B)$. However, this is not a fixed requirement, and $C$ could be either a subset or superset of it (or neither). We might deliberately choose to use some alternate conditioning set $C$, \eg, a subset of the Markov blanket to gain a more computationally efficient proposal (with a smaller proposal network), or a superset with the idea of learning longer-range structure. More fundamentally, however, Markov blankets depend on the larger graph structure might not be consistent across instantiations of a given motif (\eg, if one instantiation has additional edges connecting $B_i$ to other model variables not in $C_i$). Allowing $C$ to represent a generic conditioning set leaves us with greater flexibility in instantiating motifs.


Formally, our goal is to learn a Gibbs-like block proposal $q(B_i | C_i; \Psi_i)$ for all possible instantiations $(B_i, C_i, \Psi_i)$ of a structural motif that is close to the true conditional in the sense that \vspace{-0.1cm} \begin{equation} \label{eq:closeness}
    \forall (B_i, C_i, \Psi_i),\ \forall c_i \in \supp(C_i), q(B_i; c_i, \Psi_i) \approx p_{\Psi_i}(B_i | C_i = c_i).
\end{equation}
This provides another view of this approximation problem. If we choose the motif to have complex structures in each instantiation, the conditionals $p_{\Psi_i}(B_i | C_i = c_i)$ can often be quite different for different instantiations, and thus difficult to approximate. Therefore, choosing what is a structural motif represents a trade-off between generality of the proposal and easiness to approximate. While our approach works for any structural motif complying with the above definition, we suggest using common structures as motifs, such as chain of certain length as in Fig.~\ref{fig:motif-example}. In principle, recurring motifs could be automatically detected, but in this work, we focus on hand-identified common structures.

\subsection{Parametrizing Neural Block Proposals}\label{sec:param}

We choose mixture density networks (MDN) \citep{bishop1994mixture} as our proposal network parametrization. An MDN is a form of neural network whose outputs parametrize a mixture distribution, where in each mixture component the variables are uncorrelated. 

In our case, a neural block proposal is a function $q_\theta$ parametrized by a MDN with weights $\theta$. The function $q_\theta$ represents proposals for a structural motif $(B, C)$ by taking in current values of $C_i$ and local parameters $\Psi_i$, and outputting a distribution over $B_i$. The goal is to optimize $\theta$ so that $q_\theta$ is close to the true conditional.

In the network output, mixture weights are represented explicitly. Within each mixture component, distributions of bounded discrete variables are directly represented as independent categorical probabilities, and distributions of continuous variables are represented as isotropic Gaussians with mean and variance. To avoid degenerate proposals, we threshold the variance of each Gaussian component to be at least $10^{-5}$.

\subsection{Training Neural Block Proposals}\label{sec:neural-train}

\textbf{Loss function for a specific instantiation: }
Given a particular motif instantiation, we use the KL divergence $\kl{p_{\Psi_i}(B_i | C_i)}{q_{\theta}(B_i; C_i, \Psi_i)}$ as the measure of closeness between our proposal and the true conditional in Eq.~\ref{eq:closeness}. Taking into account all possible values $c_i \in \supp(C_i)$,  we consider the expected divergence over $C_i$'s prior:
\begin{equation}
    \mathbb{E}_{C_i}[\kl{p_{\Psi_i}(B_i | C_i)}{q_{\theta}(B_i; C_i, \Psi_i)}]  = -\mathbb{E}_{B_i, C_i}[\log q_{\theta}(B_i; C_i, \Psi_i)] + \text{constant}.
\end{equation}

The second term is independent of $\theta$. So we define the loss function on $(B_i, C_i, \Psi_i)$ as%
\begin{equation*}
    \tilde{L}(\theta; B_i, C_i, \Psi_i) = -\mathbb{E}_{B_i, C_i}[\log q_{\theta}(B_i; C_i, \Psi_i)].
\end{equation*}

\textbf{Meta-training over many instantiations: }  
To train a generalizable neural block proposal, we generate a set of random instantiations and optimize the loss function over all of them.
Assuming a distribution over instantiations $\mathcal{P}$, our goal is to minimize the overall loss
\begin{equation}
    L(\theta) = \mathbb{E}_{(B_i, C_i, \Psi_i) \sim \mathcal{P}}[\tilde{L}(\theta; B_i, C_i, \Psi_i)] = -\mathbb{E}_{(B_i, C_i, \Psi_i) \sim \mathcal{P}}[\mathbb{E}_{B_i, C_i}\left[\log q_{\theta}(B_i; C_i, \Psi_i)]\right],\label{eq:overall_loss}
\end{equation}
which is optimized with minibatch SGD in our experiments.
\hide{
\todo{Yi: I think this para is not necessary. Mentioning we can use SGD to optimize (3) is enough. If we need to cut content. remove this.}
We optimize our neural block proposals for this objective using minibatch SGD. For each training sample, we randomly sample a motif instantiation from $\mathcal{P}$, and generate samples from the prior of that instantiation. Concretely, the loss for each minibatch is obtained by
\begin{align}
    (B_i, C_i, \Psi_i) & \sim \mathcal{P} \label{eq:est-l-choose-i} \\[1pt]
    b, c & \sim p_{\Psi_i}(B_{i}, C_{i})\\[-2pt]
    \hat{L}(\theta) & = - \frac{1}{K} \sum_{j = 1}^{K} \log q_{\theta}(b; c, \Psi_{i_j})\label{eq:est-l}
\end{align}}

There are different ways to design the motif instantiation distribution $\mathcal{P}$. One approach is to find a distribution over model parameter space, and attach the random parametrizations $\Psi_i$ to $(B_i, C_i)$.\hide{, and then extract $\Psi_i$.} Practically, it is also viable to find a training dataset of models that contains a large number of instantiations. Both approaches are discussed in detail and experimented in the experiment section.




\newlength{\textfloatsepsave} \setlength{\textfloatsepsave}{\textfloatsep}%
\setlength{\textfloatsep}{6pt}%
\begin{algorithm}[t]
    \small
    \caption{\small Neural Block Sampling}
    \label{alg:mcmc}
    \begin{algorithmic}[1]
        \Require {Graphical model $(G, \Psi)$, observations $y$,\par\hspace{-0.7em} motifs~$\{(B^{(m)}, C^{(m)})\}_m$, and their instantiations $\{(B^{(m)}_i, C^{(m)}_i, \Psi^{(m)}_i)\}_{i, m}$ detected in $(G, \Psi)$.}
        \For {\textbf{each} motif $B^{(m)}, C^{(m)}$}
            \If {proposal trained for this motif exists}
                \State $q^{(m)} \longleftarrow$ trained neural block proposal
            \Else
                \State Train neural block proposal $q^{(m)}_\theta$ using SGD by Eq.~\ref{eq:overall_loss} on its instantiations $\{(B^{(m)}_i, C^{(m)}_i, \Psi^{(m)}_i)\}_i$
            \EndIf
        \EndFor
        \State $x \longleftarrow $ initialize state
        \For {$\text{timestep}$ \textbf{in} $1 \dots T$}
            \State Propose $x' \gets$ proposal $q^{(m)}_\theta$ on some instantiation $(B^{(m)}_i, C^{(m)}_i, \Psi^{(m)}_i)$
            \State Accept or reject according to MH rule
        \EndFor \\
        \Return MCMC samples
    \end{algorithmic}
\end{algorithm}%

\textbf {Neural block sampling: }
The overall MCMC sampling procedure with meta-proposals is outlined in Algorithm~\ref{alg:mcmc}, which supports building a library of neural block proposals trained on common motifs to speed up inference on previously unseen models.

\section{Experiments}\label{sec:expr}

In this section, we evaluate our method of learning neural block proposals against single-site Gibbs sampler as well as several model-specific MCMC methods. We focus on three most common structural motifs: grids, mixtures and chains. In all experiments, we use the following guideline to design the proposal: (1) using small underlying MDNs (we pick networks with two hidden layers and elu activation~\citep{clevert2015fast}), and (2) choosing an appropriate distribution to generate parameters of the motif such that the generated parameters could cover the whole space as much as possible. More experiments details and an additional experiment are available in the supplementary materials.

\setlength{\textfloatsep}{\textfloatsepsave}%
\subsection{Grid Models}\label{sec:grids}%

We start with a common structural motif in graphical models, grids. In this section, we focus on binary-valued grid models of all sorts for their relative easiness to directly compute posteriors. To evaluate MCMC algorithms, we compare the estimated posterior marginals $\hat{P}$ against true posterior marginals $P$ computed using IJGP \citep{mateescu2010join}. For each inference task with $N$ variables, we calculated the error $\frac{1}{N} \sum_{i = 1}^{N} \left\lvert \hat{P}(X_i = 1) - P(X_i = 1) \right\rvert$ as the mean absolute deviation of marginal probabilities.

\subsubsection{General Binary-Valued Grid Models}\label{sec:general-grid}%

\begin{figure}[!t]
    \centering
    \minipage[t]{0.3\textwidth}
        \centering
        \scalebox{0.55}{
            \begin{tikzpicture}[bayes_net, node distance = 0.5cm]
                \node[circle, rounded corners, minimum size = 0.35cm] (01) {};
                \node[main_node, minimum size = 0.35cm, fill=black!20] (11) [below left = of 01] {};
                \node[main_node, minimum size = 0.35cm, fill=black!20] (12) [below right = of 01] {};
                \node[main_node, minimum size = 0.35cm, fill=black!20] (21) [below left = of 11] {};
                \node[main_node, minimum size = 0.35cm] (22) [below right = of 11] {};
                \node[main_node, minimum size = 0.35cm, fill=black!20] (23) [below right = of 12] {};
                \node[main_node, minimum size = 0.35cm, fill=black!20] (31) [below left = of 21] {};
                \node[main_node, minimum size = 0.35cm] (32) [below left = of 22] {};
                \node[main_node, minimum size = 0.35cm] (33) [below left = of 23] {};
                \node[main_node, minimum size = 0.35cm, fill=black!20] (34) [below right = of 23] {};
                \node[main_node, minimum size = 0.35cm, fill=black!20] (41) [below left = of 31] {};
                \node[main_node, minimum size = 0.35cm] (42) [below left = of 32] {};
                \node[main_node, minimum size = 0.35cm] (43) [below left = of 33] {};
                \node[main_node, minimum size = 0.35cm] (44) [below left = of 34] {};
                \node[main_node, minimum size = 0.35cm, fill=black!20] (45) [below right = of 34] {};
                \node[main_node, minimum size = 0.35cm, fill=black!20] (51) [below right = of 41] {};
                \node[main_node, minimum size = 0.35cm] (52) [below right = of 42] {};
                \node[main_node, minimum size = 0.35cm] (53) [below right = of 43] {};
                \node[main_node, minimum size = 0.35cm, fill=black!20] (54) [below right = of 44] {};
                \node[main_node, minimum size = 0.35cm, fill=black!20] (61) [below right = of 51] {};
                \node[main_node, minimum size = 0.35cm] (62) [below right = of 52] {};
                \node[main_node, minimum size = 0.35cm, fill=black!20] (63) [below right = of 53] {};
                \node[main_node, minimum size = 0.35cm, fill=black!20] (71) [below right = of 61] {};
                \node[main_node, minimum size = 0.35cm, fill=black!20] (72) [below right = of 62] {};
                \node[circle, rounded corners, minimum size = 0.35cm] (00) [above left = of 11] {};
                \node[circle, rounded corners, minimum size = 0.35cm] (0E) [above right = of 12] {};
                \node[circle, rounded corners, minimum size = 0.35cm] (10) [above left = of 21] {};
                \node[circle, rounded corners, minimum size = 0.35cm] (1E) [above right = of 23] {};
                \node[circle, rounded corners, minimum size = 0.35cm] (20) [above left = of 31] {};
                \node[circle, rounded corners, minimum size = 0.35cm] (2E) [above right = of 34] {};
                \node[circle, rounded corners, minimum size = 0.35cm] (30) [above left = of 41] {};
                \node[circle, rounded corners, minimum size = 0.35cm] (3E) [above right = of 45] {};
                \node[circle, rounded corners, minimum size = 0.35cm] (50) [below left = of 41] {};
                \node[circle, rounded corners, minimum size = 0.35cm] (5E) [below right = of 45] {};
                \node[circle, rounded corners, minimum size = 0.35cm] (60) [below left = of 51] {};
                \node[circle, rounded corners, minimum size = 0.35cm] (6E) [below right = of 54] {};
                \node[circle, rounded corners, minimum size = 0.35cm] (70) [below left = of 61] {};
                \node[circle, rounded corners, minimum size = 0.35cm] (7E) [below right = of 63] {};
                \node[circle, rounded corners, minimum size = 0.35cm] (80) [below left = of 71] {};
                \node[circle, rounded corners, minimum size = 0.35cm] (81) [below right = of 71] {};
                \node[circle, rounded corners, minimum size = 0.35cm] (8E) [below right = of 72] {};

                \path[every node/.style = {font = \sffamily\small}]
                (11) edge [right] node [left] {} (21)
                (11) edge [right] node [left] {} (22)
                (12) edge [right] node [left] {} (22)
                (12) edge [right] node [left] {} (23)
                (21) edge [right] node [left] {} (31)
                (21) edge [right] node [left] {} (32)
                (22) edge [right] node [left] {} (32)
                (22) edge [right] node [left] {} (33)
                (23) edge [right] node [left] {} (33)
                (23) edge [right] node [left] {} (34)
                (31) edge [right] node [left] {} (41)
                (31) edge [right] node [left] {} (42)
                (32) edge [right] node [left] {} (42)
                (32) edge [right] node [left] {} (43)
                (33) edge [right] node [left] {} (43)
                (33) edge [right] node [left] {} (44)
                (34) edge [right] node [left] {} (44)
                (34) edge [right] node [left] {} (45)
                (41) edge [right] node [left] {} (51)
                (42) edge [right] node [left] {} (51)
                (42) edge [right] node [left] {} (52)
                (43) edge [right] node [left] {} (52)
                (43) edge [right] node [left] {} (53)
                (44) edge [right] node [left] {} (53)
                (44) edge [right] node [left] {} (54)
                (45) edge [right] node [left] {} (54)
                (51) edge [right] node [left] {} (61)
                (52) edge [right] node [left] {} (61)
                (52) edge [right] node [left] {} (62)
                (53) edge [right] node [left] {} (62)
                (53) edge [right] node [left] {} (63)
                (54) edge [right] node [left] {} (63)
                (61) edge [right] node [left] {} (71)
                (62) edge [right] node [left] {} (71)
                (62) edge [right] node [left] {} (72)
                (63) edge [right] node [left] {} (72)
                (00) edge [right, dashed, gray] node [left] {} (11)
                (01) edge [right, dashed, gray] node [left] {} (11)
                (01) edge [right, dashed, gray] node [left] {} (12)
                (0E) edge [right, dashed, gray] node [left] {} (12)
                (10) edge [right, dashed, gray] node [left] {} (21)
                (1E) edge [right, dashed, gray] node [left] {} (23)
                (20) edge [right, dashed, gray] node [left] {} (31)
                (2E) edge [right, dashed, gray] node [left] {} (34)
                (30) edge [right, dashed, gray] node [left] {} (41)
                (3E) edge [right, dashed, gray] node [left] {} (45)
                (41) edge [right, dashed, gray] node [left] {} (50)
                (45) edge [right, dashed, gray] node [left] {} (5E)
                (51) edge [right, dashed, gray] node [left] {} (60)
                (54) edge [right, dashed, gray] node [left] {} (6E)
                (61) edge [right, dashed, gray] node [left] {} (70)
                (63) edge [right, dashed, gray] node [left] {} (7E)
                (71) edge [right, dashed, gray] node [left] {} (80)
                (71) edge [right, dashed, gray] node [left] {} (81)
                (72) edge [right, dashed, gray] node [left] {} (81)
                (72) edge [right, dashed, gray] node [left] {} (8E);
            \end{tikzpicture}
        }
        \caption{Motif for general grid models. Conditioning variables (shaded) form the Markov blanket of proposed variables (white). Dashed gray arrows show possible but irrelevant dependencies\hide{ to the conditional of interest}.}
        \label{fig:general-grid-prop}
    \endminipage\hfill%
    \minipage[t]{0.68\textwidth}
        \centering
        \hspace{-0.85em}%
        \includegraphics[scale = 0.232, trim = 5 5 8 5, clip]{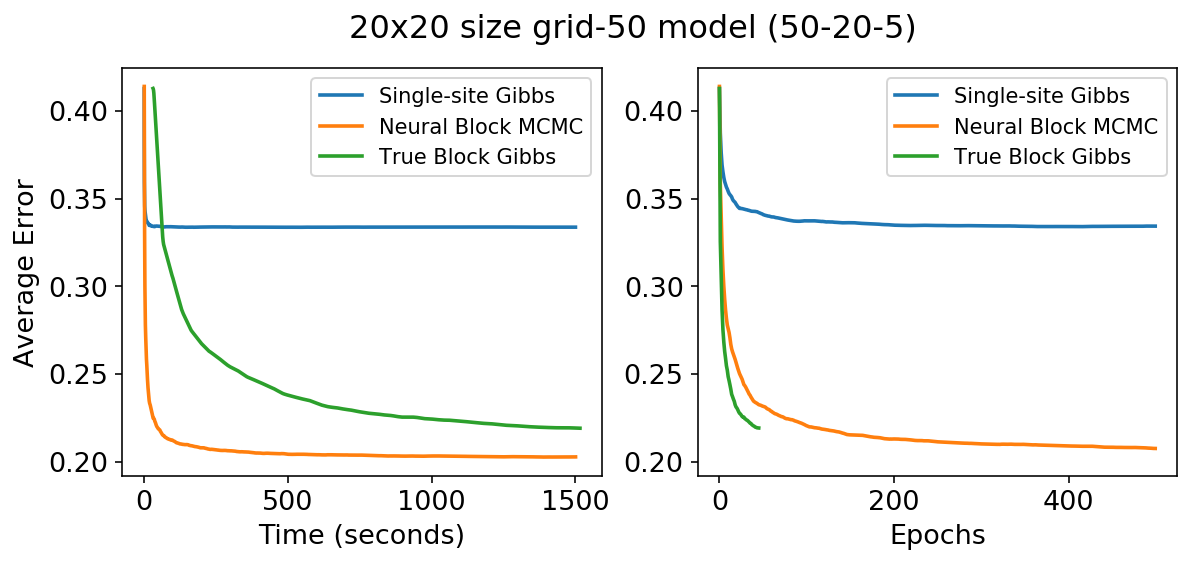}
        \includegraphics[scale = 0.232, trim = 25 5 8 0, clip]{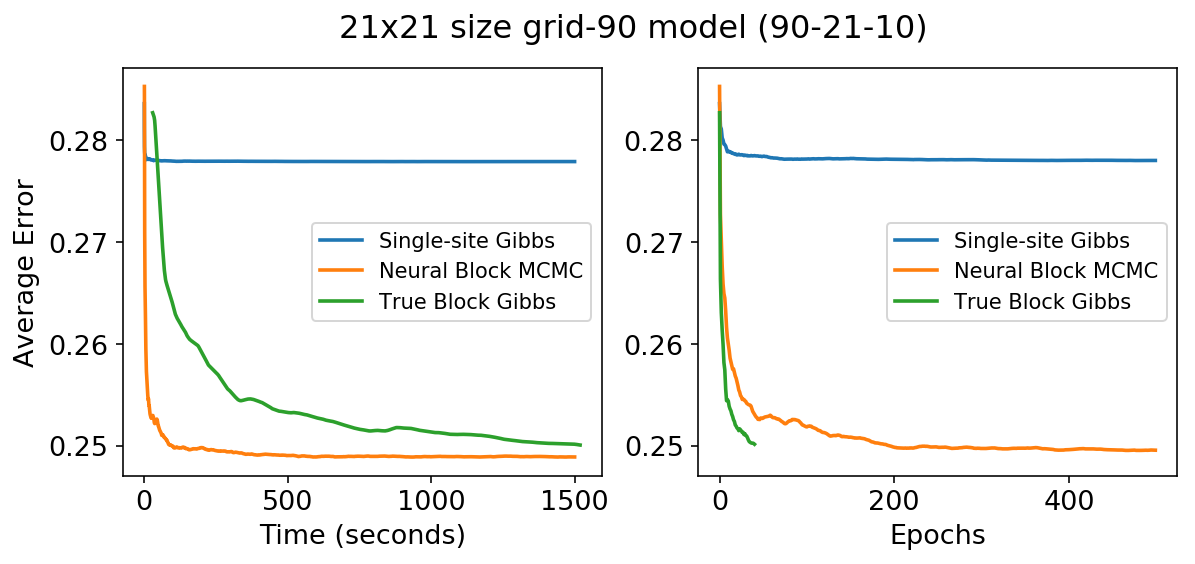}
        \caption{Sample runs comparing single-site Gibbs, Neural Block Sampling, and block Gibbs with true conditionals. \hide{Due to MCMC's sensitiveness to initialization, }For each model, we compute $10$ random initializations and run three algorithms for $1500$s on each. Epochs plots are cut off at $500$ epochs to better show the comparison because true block Gibbs finishes far less epochs within given time. \texttt{50-20-5} and \texttt{90-21-10} are identifiers of these two models in the competition.}
        \label{fig:grid-all-sample}
    \endminipage
    \vspace{1em}
    \minipage[b]{0.5\linewidth}
        \centering
        \hspace{-1em}\includegraphics[scale = 0.35, trim = 5 5 0 0, clip]{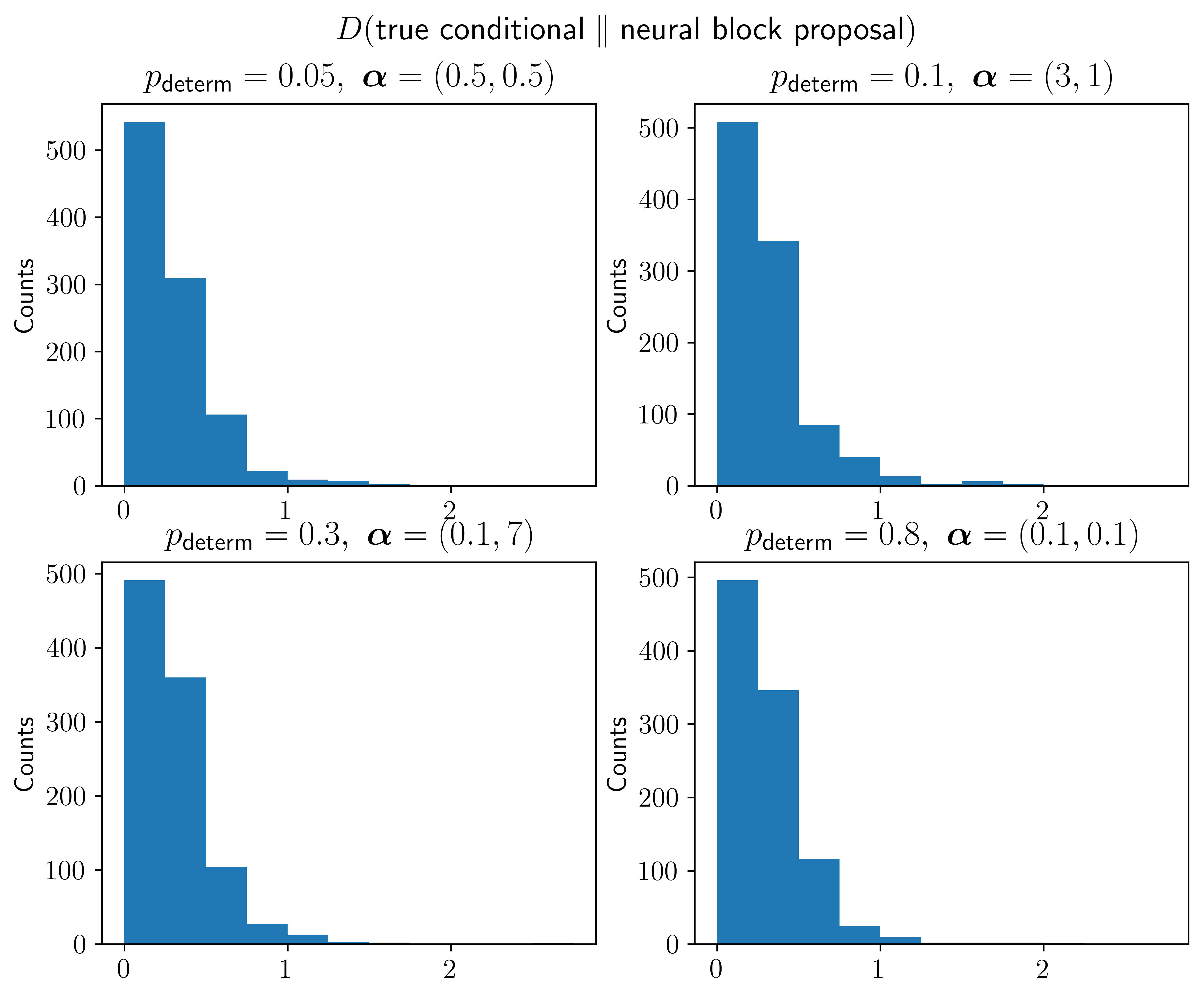}
        \caption{KL divergences between the true conditionals and our proposal outputs on $1000$ sampled instantiations from $4$ distributions with different $p_\text{determ}$ and $\boldsymbol{\alpha}$. Top left is the distribution used in training. Our trained proposal is able to generalize on arbitrary binary grid models.}
        \label{fig:grid-kl}
    \endminipage\hfill%
    \minipage[b]{0.48\linewidth}
        \centering
        \includegraphics[scale = 0.435, trim = 5 5 5 5, clip]{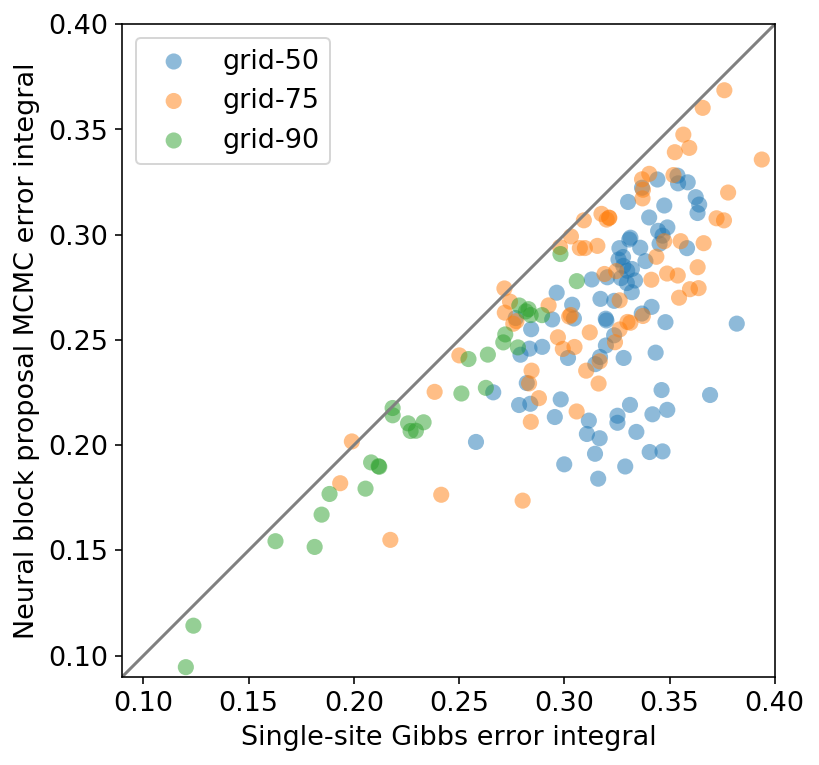}
        \caption{Performance comparison on 180 grid models from \textrm{UAI 2008 inference competition}. Each mark represents error integrals for both single-site Gibbs and our method in a single run over 1200s inference.\newline}
        \label{fig:grid-all}
    \endminipage
\end{figure}

We consider the motif in Fig.~\ref{fig:general-grid-prop}, which is instantiated in every binary-valued grid Bayesian networks (BN). Our proposal takes in the conditional probability tables (CPTs) of all $23$ variables as well as the current values of $14$ conditioning variables, and outputs a distribution over the $9$ proposed variables. 

To sample over all possible binary-valued grid instantiations, we generate random grids by sampling each CPT entry i.i.d.\ from a mixed distribution of this following form: 
\begin{equation}
    \begin{cases}
        [0, 1] & \text{w.p. } \frac{p_{\text{determ}}}{2} \\
        [1, 0] & \text{w.p. } \frac{p_{\text{determ}}}{2} \\
        \text{Dirichlet}(\boldsymbol{\alpha}) & \text{w.p. } 1 - p_{\text{determ}}, \\
    \end{cases}\label{eq:bn-distn}
\end{equation}
where $p_{\text{determ}} \in [0, 1]$ is the probability of the CPT entry being deterministic. Our proposal is trained with $p_\text{determ} = 0.05$ and $\boldsymbol{\alpha} = (0.5, 0.5)$.

To test the generalizability of our trained proposal, we generate random binary grid instantiations using distributions with various $p_\text{determ}$ and $\boldsymbol{\alpha}$ values, and compute the KL divergences between the true conditionals and our proposal outputs on $1000$ sampled instantiations from each distribution. Fig.~\ref{fig:grid-kl} shows the histograms of divergence values from $4$ very different distributions, including the one used for training (top left). The resulting histograms show mostly small divergence values, and are nearly indistinguishable, even though one distribution has $p_\text{determ} = 0.8$ and the proposal is only trained with $p_\text{determ} = 0.05$. This shows that our approach is able to generally and accurately approximate true conditionals, despite only being trained with an arbitrary distribution.

We evaluate the performance of the trained neural block proposal on all $180$ grid BNs up to $500$ nodes from \textrm{UAI 2008}\hide{\footnote{\url{http://graphmod.ics.uci.edu/uai08}}} inference competition. In each epoch, for each latent variable, we try to identify and propose the block as in Fig.~\ref{fig:general-grid-prop} with the variable located at center. If this is not possible, \eg, the variable is at boundaries or close to evidence, single-site Gibbs resampling is used instead.

Fig.~\ref{fig:grid-all} shows the performance of both our method and single-site Gibbs in terms of error integrated over time for all $180$ models. The models are divided into three classes, grid-$50$, grid-$75$ and grid-$90$, according to the percentage of deterministic relations. Our neural block sampler significantly outperforms Gibbs sampler in nearly every model. We notice that the improvement is less significant as the percentage of deterministic relations increases. This is largely due to that the above proposal structure in Fig.~\ref{fig:general-grid-prop} can only easily handle dependency among the $9$ proposed nodes. We expect an increased block size to yield stronger performance on models with many deterministic relations.

Furthermore, we compare our proposal against single-site Gibbs, and exact block Gibbs with identical proposal block, on grid models with different percentages of deterministic relations in Fig.~\ref{fig:grid-all-sample}. Single-site Gibbs performs worst on both models due to quickly getting stuck in local modes. Between the two block proposals, neural block sampling performs better in error w.r.t.~time due to shorter computational time. However, because the neural block proposal is only an approximate of the true block Gibbs proposal, it is worse in terms of error w.r.t.~epochs, as expected. Detailed comparisons on more models are available in the supplementary material.

\hide{In summary, our experiment results show that neural block proposals can achieve significantly faster and better mixing than single-site Gibbs, while using much less computation overhead than computing the exact Gibbs block proposal.}

Additionally, our approach can be used model-specifically by training only on instantiations within a particular model. In supplementary materials, we demonstrate that our method achieves comparable performance with a more advanced task-specific MCMC method, Inverse MCMC~\citep{stuhlmuller2013learning}.

\subsection{Gaussian Mixture Model with Unknown Number of Components}
\label{sec:gmm}
We next consider open-universe Gaussian mixture models (GMMs), in which the number of mixture components is unknown, subject to a prior. Similarly to Dirichlet process GMMs, these are typically treated with hand-designed model-specific split-merge MCMC algorithms.%
\begin{figure}[!t]
    \centering
    \hspace{0.05cm}
    \minipage[t]{0.49\textwidth}
        \centering
        \hspace{-0.9em}\includegraphics[scale = 0.32, trim = 5 6 0 0, clip]{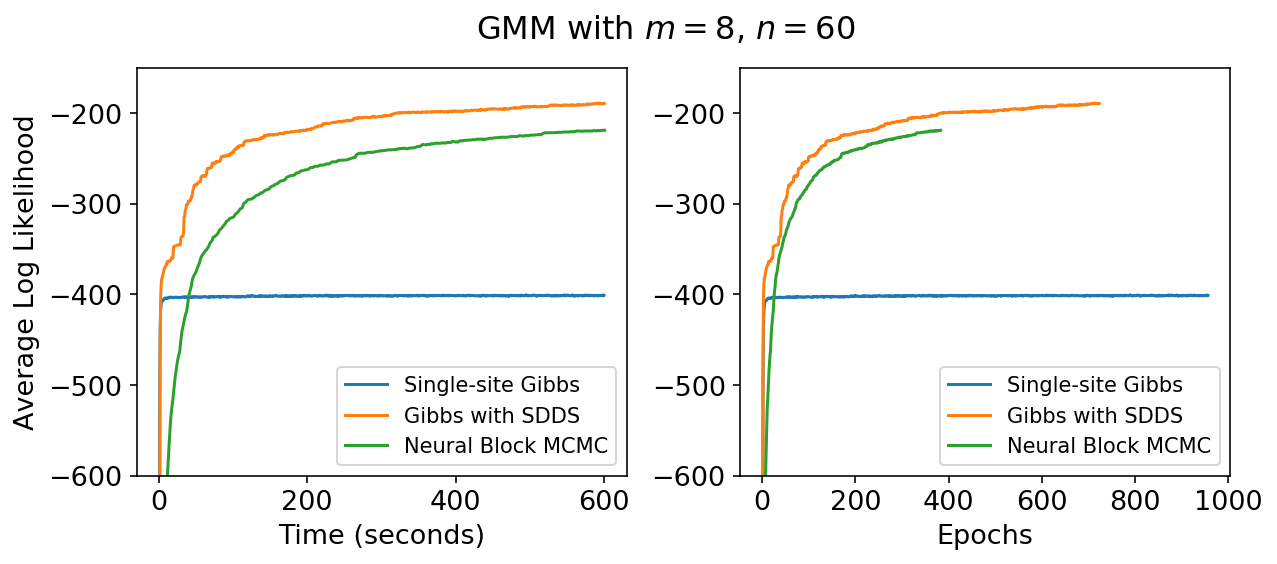} \\
        \vspace{0.18em}\vspace{1pt}
        \hspace{-1.24em}\includegraphics[scale = 0.32, trim = 5 6 0 0, clip]{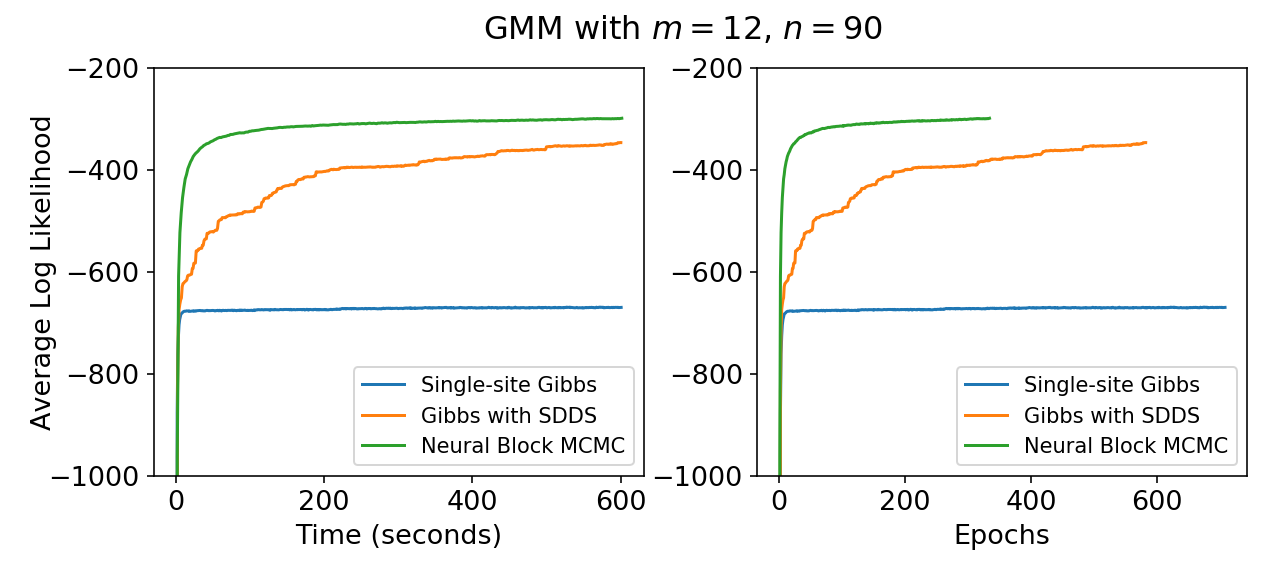}
    \endminipage\hfill%
    \minipage[t]{0.49\textwidth}
        \centering
        \hspace{-1.1em}\includegraphics[scale = 0.32, trim = 5 6 0 0, clip]{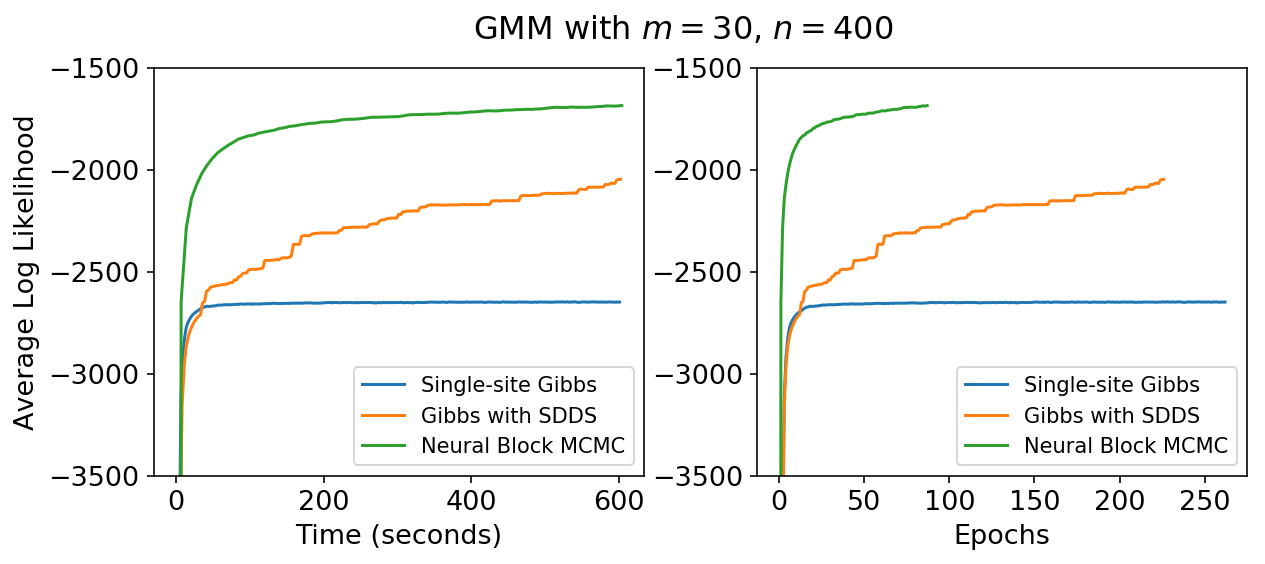} \\
        \vspace{1pt}
        \hspace{-0.2em}\includegraphics[scale = 0.32, trim = 5 7 0 6, clip]{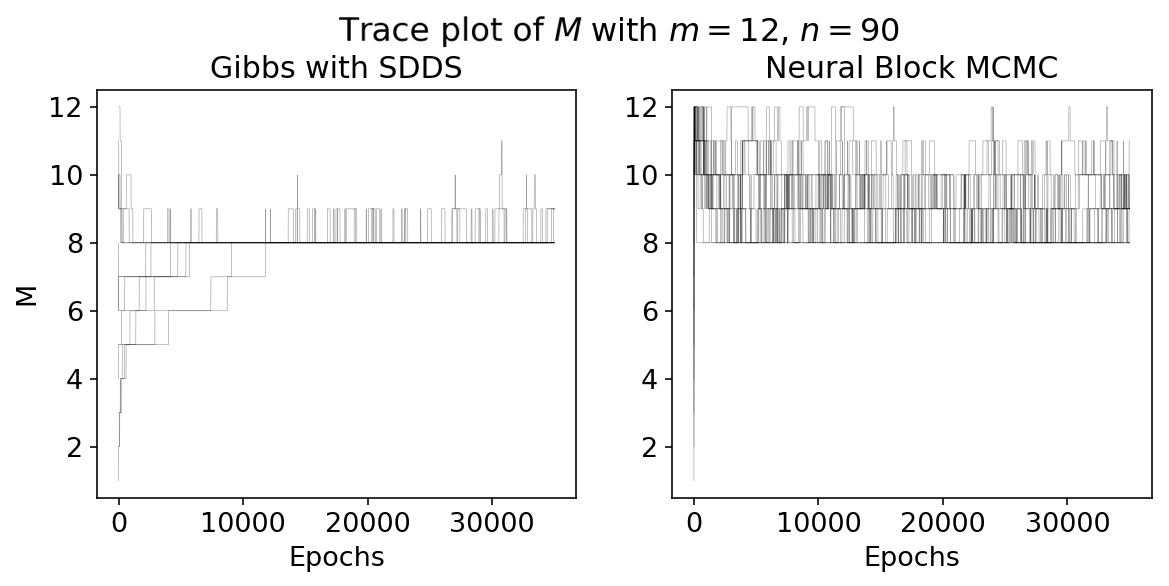}
    \endminipage
    \caption{\textbf{All except bottom right: }Average log likelihoods of MCMC runs over $200$ tasks for total $600$s in various GMMs. \textbf{Bottom right: }Trace plots of $M$ over $12$ runs from initialization with different $M$ values on a GMM with $m = 12$, $n = 90$. Our approach explores sample space much faster than Gibbs with SDDS.}
    \label{fig:gmm-ll}
    \label{fig:gmm-trace}
\end{figure}

Consider \hide{a GMM in Fig.~\ref{fig:gmm}}the following GMM. $n$ points $\mathbf{x} = \{x_i\}_{i = 1, \dots, n}$ are observed, and come uniformly randomly from one of $M$ (unknown) active mixtures, with $M \sim \text{Unif}\{1, 2, \dots, m\}$. Our task is to infer the posterior of mixture means $\boldsymbol{\mu} = \{\mu_j\}_{j = 1, \dots, M}$, their activity indicators $\mathbf{v} = \{v_j\}_{j = 1, \dots, M}$, and the labels $\mathbf{z} = \{z_i\}_{i = 1, \dots, n}$, where $z_i$ is the mixture index $x_i$ comes from. Since $M$ is determined by $\mathbf{v}$, in this experiment, we always calculate $M = \sum_j v_j$ instead of sampling $M$.

Such GMMs have many nearly-deterministic relations, \eg, $p(v_j = 0, z_i = j) = 0$, causing vanilla single-site Gibbs failing to jump across different $M$ values. Split-merge MCMC algorithms, \eg, Restricted Gibbs split-merge (RGSM) \citep{jain2004split} and Smart-Dumb/Dumb-Smart (SDDS) \citep{wang2015smart}, use hand-designed MCMC moves to solve such issues.
In our framework, it's possible to deal with such relations with a proposal block including all of $\mathbf{z}$, $\boldsymbol{\mu}$ and $\mathbf{v}$. However, doing so requires significant training and inference time (due to larger proposal network and larger proposal block), and the resulting proposal can not generalize to GMMs of different sizes. 
%

In order to apply the trained proposal to differently sized GMMs, we choose the motif to propose $q_\theta$ for two arbitrary mixtures $(\mu_i, v_i)$ and $(\mu_j, v_j)$ conditioned on all other variables {\em excluding $\mathbf{z}$}, and instead consider the model with $\mathbf{z}$ variables collapsed. The inference task is then equivalent to first sampling $\boldsymbol{\mu}, \mathbf{v}$ from the collapsed model $p(\boldsymbol{\mu}, \mathbf{v} | \mathbf{x})$, and then $\mathbf{z}$ from $p(\mathbf{z} | \boldsymbol{\mu}, \mathbf{v}, \mathbf{x})$. We modify the algorithm such that the proposal from $q_\theta$ is accepted or rejected by the MH rule on the {\em collapsed} model. Then $\mathbf{z}$ is resampled from $p(\mathbf{z} | \boldsymbol{\mu}, \mathbf{v}, \mathbf{x})$. This approach is less sensitive to different $n$ values and performs well in variously sized GMMs. More details are available in the supplementary material.

\hide{\endgroup}

We train with a small GMM with $m = 8$ and $n = 60$ as the motif, and apply the trained proposal on GMMs with larger $m$ and $n$ by randomly selecting $8$ mixtures and $60$ points for each proposal. Fig.~\ref{fig:gmm-ll} shows how the our sampler performs on GMM of various sizes, compared against split-merge Gibbs with SDDS. We notice that as model gets larger, Gibbs with SDDS mixes more slowly, while neural block sampling still mixes fairly fast and outperforms Gibbs with SDDS. Bottom right of Fig.~\ref{fig:gmm-trace} shows the trace plots of $M$ for both algorithms over multiple runs on the same observations. Gibbs with SDDS takes a long time to find a high likelihood explanation and fails to explore other possible ones efficiently. Our proposal, on the other hand, mixes quickly among the possible explanations.

\subsection{Named Entity Recognition (NER) Tagging}\label{sec:ner}
\begin{figure}
    \centering
    \hspace{-0.8em}\includegraphics[scale = 0.335, trim = 5 5 0 0, clip]{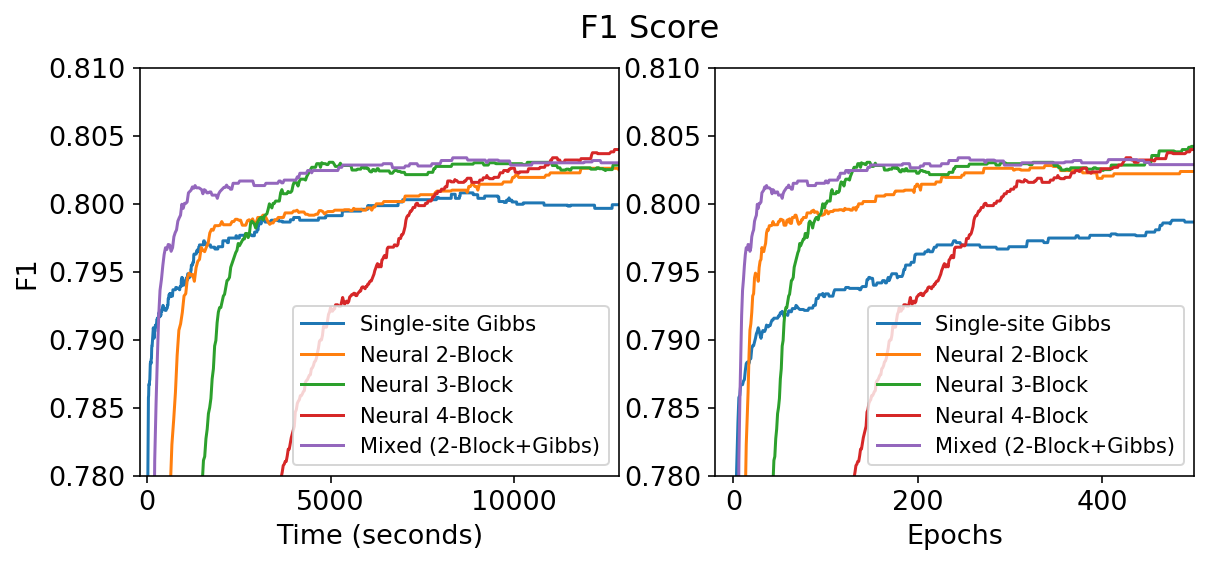}\hfill%
    \includegraphics[scale = 0.335, trim = 5 5 0 0, clip]{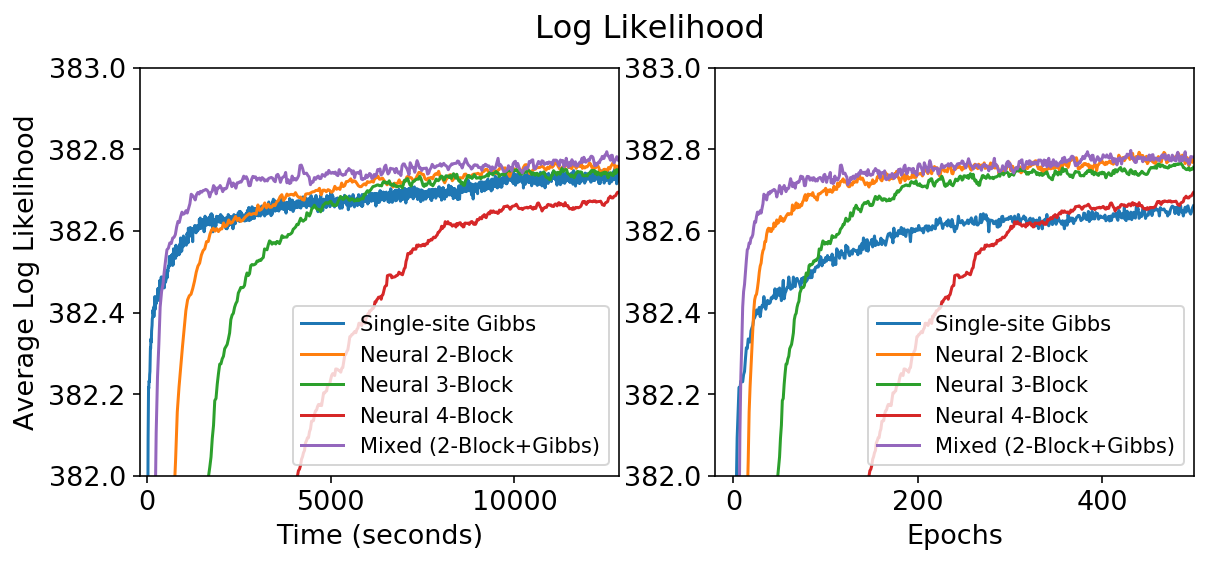}
    \caption{Average F1 scores and average log likelihoods over entire test dataset. In each epoch, all variables in every test MRF is proposed roughly once for all algorithms. F1 scores are measured using states with highest likelihood seen over Markov chain traces. To better show comparison, epoch plots are cut off at $500$ epochs and time plots at $12850$s. Log likelihoods shown don't include normalization constant.}
    \label{fig:ner_f1_ll}
\end{figure}

Named entity recognition (NER) is the task of inferring named entity tags for words in natural language sentences. One way to tackle NER is to train a conditional random field (CRF) model representing the joint distribution of tags and word features~\citep{liang2008structure}. \hide{In particular, the model contains weights between word features and tags as well as high order factors of consecutive tags. For each test sentence, we build a chain Markov random field (MRF) containing only the tags variables using word features and learned CRF model, and apply MCMC methods like single-site Gibbs to sample the NER tags.} In test time, we use the CRF build a chain Markov random field (MRF) containing only tags variables, and apply MCMC methods to sample the NER tags. We use a dataset of $17494$ sentences from CoNLL-2003 Shared Task\footnote{\url{https://www.clips.uantwerpen.be/conll2003/ner/}}. The CRF model is trained with AdaGrad \citep{duchi2011adaptive} through $10$ sweeps over the training dataset.

Our goal is to train good neural block proposals for the chain MRFs built for test sentences. Experimenting with different chain lengths, we train three proposals, each for a motif of two, three, or four consecutive proposed tag variables and their Markov blanket. \hide{With each proposal, the MDN takes in both the local MRF parameters and assignments of Markov blanket variables, then outputs the proposal as a mixture of $4$ components. }These proposals are trained on instantiations within MRFs built from the training dataset for the CRF model.%

We then evaluate the learned neural block proposals on the previously unseen test dataset of $3453$ sentences. Fig.~\ref{fig:ner_f1_ll} plots the performance of neural block sampling and single-site Gibbs w.r.t.~both time and epochs on the entire test dataset. As block size grows larger, learned proposal takes more time to mix. But eventually, block proposals generally achieve better performance than single-site Gibbs in terms of both F1 scores and log likelihoods. Therefore, as shown in the figure, a mixed proposal of single-site Gibbs and neural block proposals can achieve better mixing without slowing down much. As an interesting observation, neural block sampling sometimes achieves higher F1 scores even before surpassing single-site Gibbs in log likelihood, implying that log likelihood is at best an imperfect proxy for performance on this task.

\section{Conclusion}\label{sec:discuss}

This paper proposes and explores the (to our knowledge) novel idea of meta-learning generalizable approximate block Gibbs proposals. Our meta-proposals are trained offline and can be applied directly to novel models given only a common set of structural motifs. Experiments show that the neural block sampling approach outperforms standard single-site Gibbs in both convergence speed and sample quality \hide{can help overcome bad local modes comparing with single-site Gibbs sampling }and achieve comparable performance against model-specialized methods. In will be an interesting system design problem to investigate, when given a library of trained block proposals, how an inference system in a probabilistic programming language can automatically detect the common structural motifs and (adaptively) apply appropriate samplers to help convergence for more general real-world applications.

Additionally, from the meta-learning perspective, our method is based on meta-training, \ie, training over a variety of motif instantiations. At test time, the learned proposal does not adapt to new scenarios after meta-training. While in many meta-learning works in reinforcement learning~\cite{finn2017model,duan2016rl}, a meta-trained agent can further adapt the learned policy to unseen environments via a few learning steps under the assumption that a reward signal is accessible at test time. In our setting, we can similarly adopt such fast adaptation scheme at test time to further improve the quality of proposed samples by treating the acceptance rate as a test time reward signal. We leave this as a future work.
\hide{
\todo{Yi: remove this para}In the current stage, our framework requires the user to manually detect common structural motifs and choose where and how to apply the pretrained block samplers. It will be a very interesting direction to investigate, when given a library of trained block proposals, how an inference system can automatically detect the common structural motifs and (adaptively) apply appropriate samplers to help convergence for more general real-world applications.}

\newpage\clearpage

\bibliography{reference}
\bibliographystyle{plainnat}

\newpage\clearpage


\setcounter{section}{0}
\renewcommand{\thesection}{S-\arabic{section}}

\section{Experiment Details}
As mentioned in main paper Sec.~\ref{sec:expr}, parametrizing MDNs in all experiments have elu activation and two hidden layers each of size $\lambda \max\left\{\text{input size}, \text{output size}\right\}$, where $4 \leq \lambda \leq 5$ depending on the task, and output the proposal distribution as a mixture of $4 \leq m \leq 16$ components.

\subsection{General Binary-Valued Grid Models}
For directed binary-valued grid models, we use higher amount of MDN mixtures than other experiments because more variables are proposed and general discrete BNs can have highly multi-modal conditionals.

\begin{figure}
    \centering
    \hspace{-1em}\includegraphics[scale = 0.39, trim = 5 5 0 0, clip]{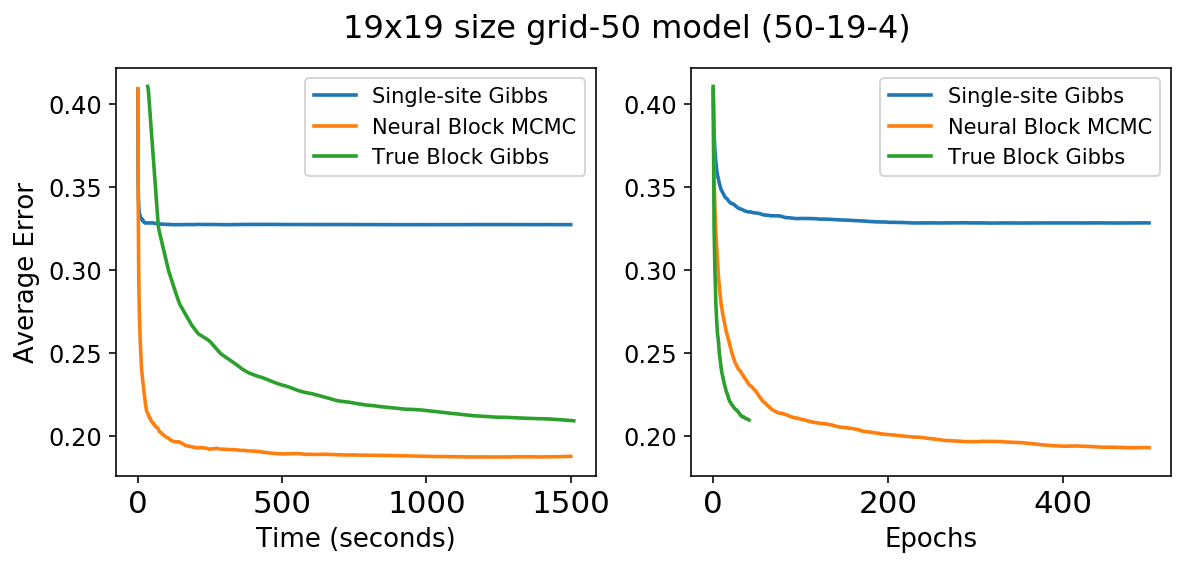}\\
    \hspace{-1em}\includegraphics[scale = 0.39, trim = 5 5 0 0, clip]{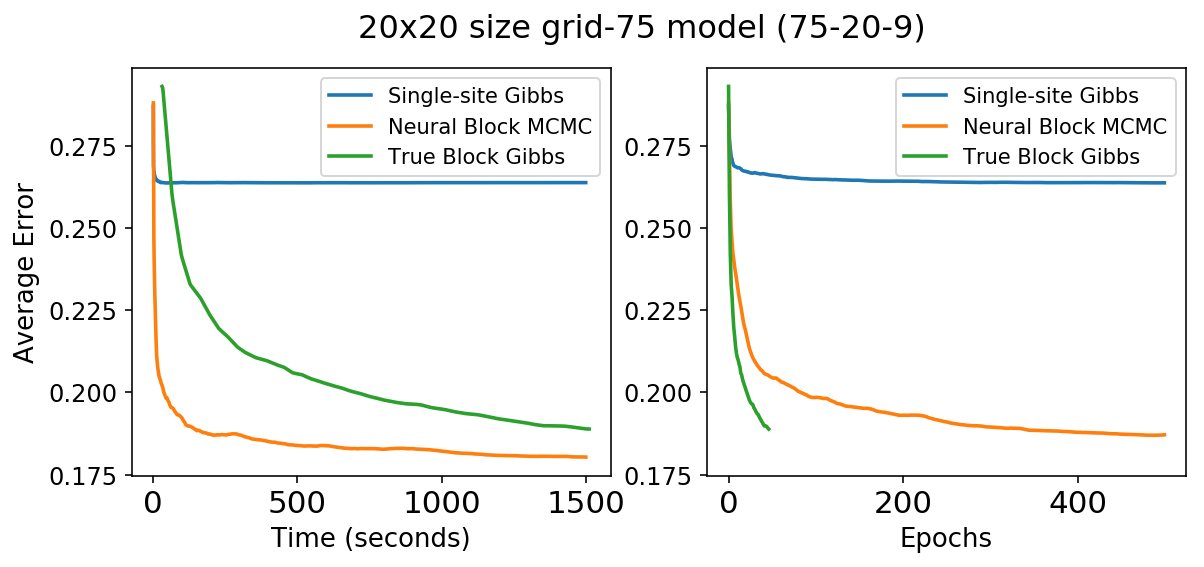}\\
    \hspace{-1em}\includegraphics[scale = 0.39, trim = 5 5 0 0, clip]{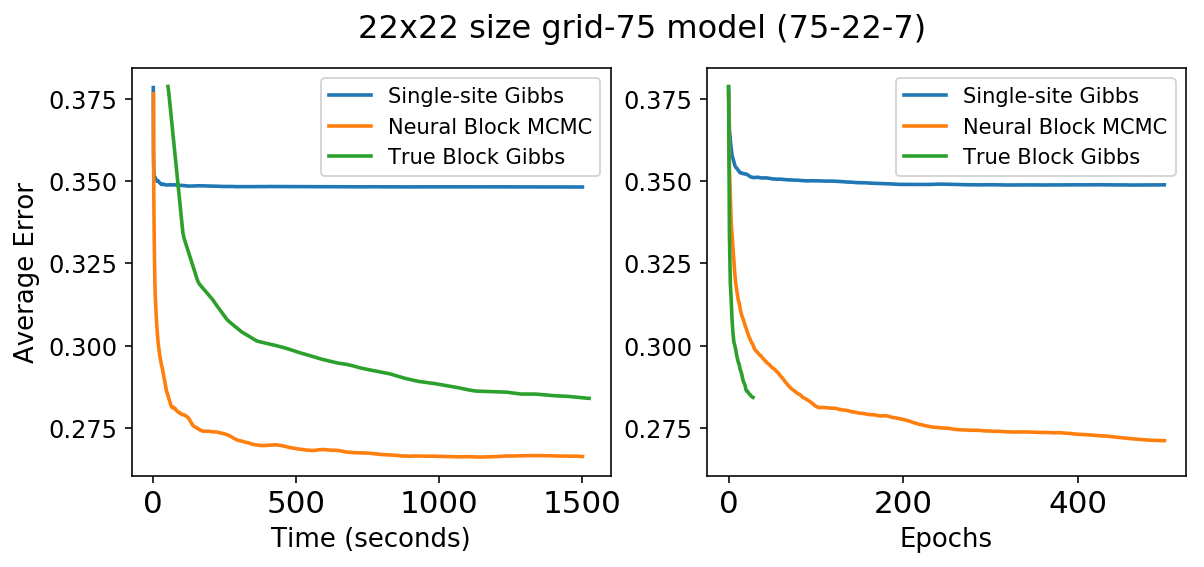}\\
    \hspace{-1em}\includegraphics[scale = 0.39, trim = 5 5 0 0, clip]{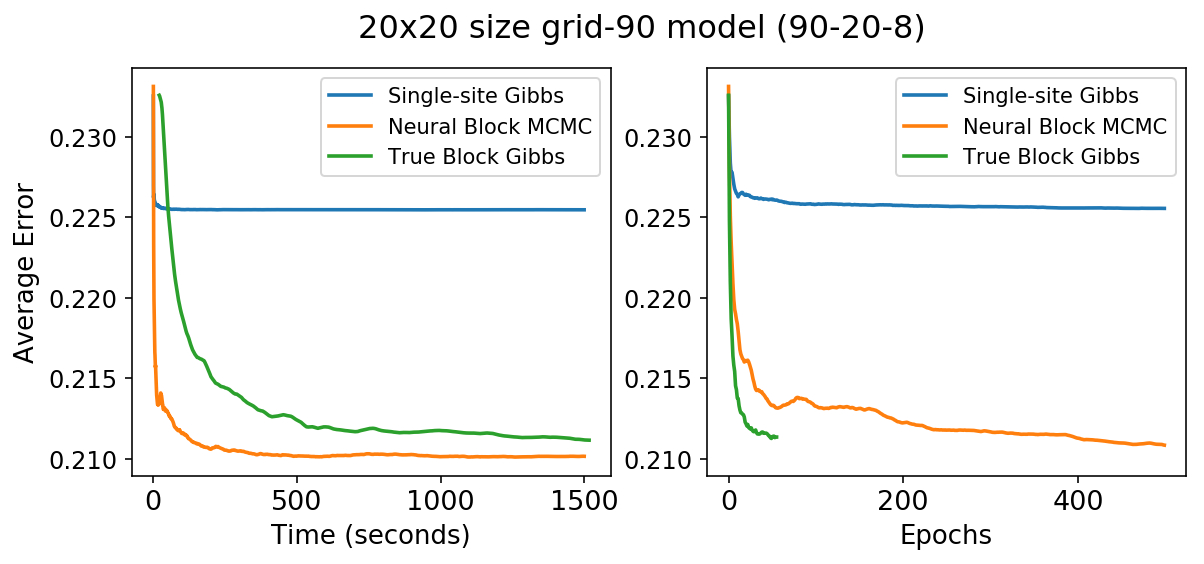}\\
    \caption{Additional sample runs with single-site Gibbs, neural block MCMC, and block Gibbs with true conditionals on UAI 2008 grid models. These results are obtained in same setting as Fig.~\ref{fig:grid-all-sample} of main paper. For each model, we compute $10$ random initializations and run three algorithms for $1500$s on each one. Plots show average error for each algorithm. Epochs plots are cut off at $500$ epochs to better show the comparison. ``grid-$k$'' represents that the model has $k\%$ deterministic relations.}
    \label{fig:grid-all-more-sample}
\end{figure}

\paragraph{Architecture}The underlying MDN has 106-480-480-120 network structure, mapping the CPTs of all $23$ motif variables and $14$ conditioning variable values to the proposal distribution of $9$ proposed variables as a mixture of $12$ components.
\paragraph{Additional Sample Runs} We provide additional sample runs on four grid models with various percentages of deterministic relations in Fig.~\ref{fig:grid-all-more-sample}.
\subsection{Gaussian Mixture Model with Unknown Number of Components}

\paragraph{Formal Model} Formally, the model can be written as
\begin{align*}
    M & \sim \text{Unif}\{1, 2, \dots, m\} \\
    \mu_j & \sim \N(0, \sigma^2_\mu I)  & j = 1, \dots, m \\
    \mathbf{v} | M & \sim \text{Unif}\{a \in \{0, 1\}^m \colon \sum_j a_j = M\} \hspace{-2em} \\[-10pt]
    z_i | \mathbf{v} & \sim \text{Unif}\{j \colon v_j = 1\} & i = 1, \dots, n \\
    x_i | z_i, \boldsymbol{\mu} & \sim \N(\mu_{z_i}, \sigma^2 I) & i = 1, \dots, n\makebox[0pt][l]{,}
\end{align*}\\%
where $m$ and $n$ are model parameters, and $\sigma^2_\mu = 4$ and $\sigma^2 = 0.1$ are fixed constants.

\paragraph{Architecture}Our neural block proposal is trained using the GMM with $m = 8$ max mixtures and $n = 60$ data points. It proposes two mixture components $(\mu_j, v_j)$s through underlying MDN of 156-624-624-36 network structure. The MDN's inputs include $60$ observed points $\mathbf{x} = \{x_i\}_i$, $8$ component means $\boldsymbol{\mu} = \{\mu_j\}_j$ and component active indicators $\mathbf{v} = \{v_j\}_j$ with values for the two proposed mixtures replaced by zeros. Orders for $\mathbf{x}$, $\boldsymbol{\mu}$ and $\mathbf{v}$ are such that they are sorted along the first principle component of $\mathbf{x}$ to break symmetry. In addition, the inputs also contain the principle component and the indicators of which components are being proposed. The MDN outputs a proposal distribution over the two $(\mu_j, v_j)$s as a mixture of 4 MDN components.

\paragraph{Breaking the Symmetry} In training, such mixture models have symmetries that must be broken before being used as input to the neural network \citep{nishihara2013detecting}. In particular, the mixtures $\{(v_j, \mu_j)\}_j$ can be permuted in $m!$ ways and the points $\{(z_i, x_i)\}_i$ in $n!$ ways. Following a similar procedure by \citet{le2017inference}, we sort these values according the first principal component of $\mathbf{x}$, and also feed the first principal component vector into the MDN.

\paragraph{Proposal and Inference with $\mathbf{z}$ Collapsed}In order to avoid nearly-deterministic relations, \eg, $p(v_j = 0, z_i = j) = 0$, and still train a general proposal unconstrained by $n$, we choose to consider the collapsed model without $\mathbf{z}$. We first experiment on the intuitive approach which adds a resampling step for $\mathbf{z}$ in the proposal. At each proposal step, trained proposal $q_\theta$ is first used to propose new mixtures $\boldsymbol{\mu}'$ and $\mathbf{v}'$, and then $\mathbf{z}$ is proposed from $p(\mathbf{z} | \boldsymbol{\mu}', \mathbf{v}', \mathbf{x})$. The MH rule is applied lastly to either accept or reject all proposed values. While this method gives good performance in small models, it suffers greatly from low acceptance ratio as $n$, the number of observed points $\mathbf{z}$, grows large. Therefore, we eventually choose the approach described in main paper Sec.~\ref{sec:gmm}, \ie, applying the MH rule on the {\em collapsed} model with $\mathbf{z}$ resampled from $p(\mathbf{z} | \boldsymbol{\mu}, \mathbf{v}, \mathbf{x})$ afterwards. Since the acceptance ratio no longer depends on $n$, this approach behaves much more scalable than the first one in our experiments. It outperforms SDDS split-merge MCMC in GMMs of various sizes, as shown in Fig.~\ref{fig:gmm-ll} of main paper.
\subsection{NER Tagging}
\paragraph{Architecture}The underlying MDN has two hidden layers each of size $4 \times \max\left\{\text{input size}, \text{output size}\right\}$, with output size varying according to the number of proposed variables. It maps local CRF parameters of all motif variables and conditioning variable values to the NER tag proposal for consecutive proposed variables as a mixture of $4$ components.

\section{Additional Experiment}
\subsection{Comparison with Inverse MCMC}
\label{sec:expr-inverse-mcmc}

\begin{figure}[t]
    \centering
    \minipage[b]{0.275\textwidth}
        \centering
        \hspace{-0.2em}\scalebox{0.5}{
            \begin{tikzpicture}[bayes_net, node distance = 0.5cm]
                \node[main_node, minimum size = 0.35cm] (01) {};
                \node[main_node, minimum size = 0.35cm] (11) [below left = of 01] {};
                \node[main_node, minimum size = 0.35cm] (12) [below right = of 01] {};
                \node[main_node, minimum size = 0.35cm] (21) [below left = of 11] {};
                \node[main_node, minimum size = 0.35cm] (22) [below right = of 11] {};
                \node[main_node, minimum size = 0.35cm] (23) [below right = of 12] {};
                \node[main_node, minimum size = 0.35cm] (31) [below left = of 21] {};
                \node[main_node, minimum size = 0.35cm] (32) [below left = of 22] {};
                \node[main_node, minimum size = 0.35cm] (33) [below left = of 23] {};
                \node[main_node, minimum size = 0.35cm] (34) [below right = of 23] {};
                \node[main_node, minimum size = 0.35cm] (41) [below left = of 31] {};
                \node[main_node, minimum size = 0.35cm] (42) [below left = of 32] {};
                \node[main_node, minimum size = 0.35cm] (43) [below left = of 33] {};
                \node[main_node, minimum size = 0.35cm] (44) [below left = of 34] {};
                \node[main_node, minimum size = 0.35cm] (45) [below right = of 34] {};
                \node[main_node, minimum size = 0.35cm, fill=black!20] (51) [below left = of 41] {};
                \node[main_node, minimum size = 0.35cm, fill=black!20] (52) [below left = of 42] {};
                \node[main_node, minimum size = 0.35cm, fill=black!20] (53) [below left = of 43] {};
                \node[main_node, minimum size = 0.35cm, fill=black!20] (54) [below left = of 44] {};
                \node[main_node, minimum size = 0.35cm, fill=black!20] (55) [below left = of 45] {};
                \node[main_node, minimum size = 0.35cm, fill=black!20] (56) [below right = of 45] {};

                \path[every node/.style = {font = \sffamily\small}]
                (01) edge [right] node [left] {} (11)
                (01) edge [right] node [left] {} (12)
                (11) edge [right] node [left] {} (21)
                (11) edge [right] node [left] {} (22)
                (12) edge [right] node [left] {} (22)
                (12) edge [right] node [left] {} (23)
                (21) edge [right] node [left] {} (31)
                (21) edge [right] node [left] {} (32)
                (22) edge [right] node [left] {} (32)
                (22) edge [right] node [left] {} (33)
                (23) edge [right] node [left] {} (33)
                (23) edge [right] node [left] {} (34)
                (31) edge [right] node [left] {} (41)
                (31) edge [right] node [left] {} (42)
                (32) edge [right] node [left] {} (42)
                (32) edge [right] node [left] {} (43)
                (33) edge [right] node [left] {} (43)
                (33) edge [right] node [left] {} (44)
                (34) edge [right] node [left] {} (44)
                (34) edge [right] node [left] {} (45)
                (41) edge [right] node [left] {} (51)
                (41) edge [right] node [left] {} (52)
                (42) edge [right] node [left] {} (52)
                (42) edge [right] node [left] {} (53)
                (43) edge [right] node [left] {} (53)
                (43) edge [right] node [left] {} (54)
                (44) edge [right] node [left] {} (54)
                (44) edge [right] node [left] {} (55)
                (45) edge [right] node [left] {} (55)
                (45) edge [right] node [left] {} (56);
            \end{tikzpicture}
        }
        \caption{Small version of the triangle grid model in experiment Sec.~\ref{sec:stochastic_inverse_experiments}. Evidence nodes (shaded) are at bottom layer. The actual network has 15 layers and 120 nodes.}
        \label{fig:triangle-grid}
    \endminipage\hfill
    \minipage[b]{0.29\textwidth}
        \centering
        \hspace{-0.5em}\scalebox{0.5}{
            \begin{tikzpicture}[bayes_net, node distance = 0.5cm]
                \node[circle, rounded corners, minimum size = 0.35cm] (01) {};
                \node[main_node, minimum size = 0.35cm, fill=black!20] (11) [below left = of 01] {};
                \node[main_node, minimum size = 0.35cm, fill=black!20] (12) [below right = of 01] {};
                \node[main_node, minimum size = 0.35cm, fill=black!20] (21) [below left = of 11] {};
                \node[main_node, minimum size = 0.35cm] (22) [below right = of 11] {};
                \node[main_node, minimum size = 0.35cm, fill=black!20] (23) [below right = of 12] {};
                \node[main_node, minimum size = 0.35cm, fill=black!20] (31) [below left = of 21] {};
                \node[main_node, minimum size = 0.35cm] (32) [below left = of 22] {};
                \node[main_node, minimum size = 0.35cm] (33) [below left = of 23] {};
                \node[main_node, minimum size = 0.35cm, fill=black!20] (34) [below right = of 23] {};
                \node[main_node, minimum size = 0.35cm, fill=black!20] (41) [below left = of 31] {};
                \node[main_node, minimum size = 0.35cm] (42) [below left = of 32] {};
                \node[main_node, minimum size = 0.35cm] (43) [below left = of 33] {};
                \node[main_node, minimum size = 0.35cm] (44) [below left = of 34] {};
                \node[main_node, minimum size = 0.35cm, fill=black!20] (45) [below right = of 34] {};
                \node[main_node, minimum size = 0.35cm, fill=black!20] (51) [below left = of 41] {};
                \node[main_node, minimum size = 0.35cm] (52) [below left = of 42] {};
                \node[main_node, minimum size = 0.35cm] (53) [below left = of 43] {};
                \node[main_node, minimum size = 0.35cm] (54) [below left = of 44] {};
                \node[main_node, minimum size = 0.35cm] (55) [below left = of 45] {};
                \node[main_node, minimum size = 0.35cm, fill=black!20] (56) [below right = of 45] {};
                \node[main_node, minimum size = 0.35cm, fill=black!20] (61) [below right = of 51] {};
                \node[main_node, minimum size = 0.35cm] (62) [below right = of 52] {};
                \node[main_node, minimum size = 0.35cm] (63) [below right = of 53] {};
                \node[main_node, minimum size = 0.35cm] (64) [below right = of 54] {};
                \node[main_node, minimum size = 0.35cm, fill=black!20] (65) [below right = of 55] {};
                \node[main_node, minimum size = 0.35cm, fill=black!20] (71) [below right = of 61] {};
                \node[main_node, minimum size = 0.35cm, fill=black!20] (72) [below right = of 62] {};
                \node[main_node, minimum size = 0.35cm, fill=black!20] (73) [below right = of 63] {};
                \node[main_node, minimum size = 0.35cm, fill=black!20] (74) [below right = of 64] {};
                \node[circle, rounded corners, minimum size = 0.35cm] (00) [above left = of 11] {};
                \node[circle, rounded corners, minimum size = 0.35cm] (0E) [above right = of 12] {};
                \node[circle, rounded corners, minimum size = 0.35cm] (10) [above left = of 21] {};
                \node[circle, rounded corners, minimum size = 0.35cm] (1E) [above right = of 23] {};
                \node[circle, rounded corners, minimum size = 0.35cm] (20) [above left = of 31] {};
                \node[circle, rounded corners, minimum size = 0.35cm] (2E) [above right = of 34] {};
                \node[circle, rounded corners, minimum size = 0.35cm] (30) [above left = of 41] {};
                \node[circle, rounded corners, minimum size = 0.35cm] (3E) [above right = of 45] {};
                \node[circle, rounded corners, minimum size = 0.35cm] (40) [above left = of 51] {};
                \node[circle, rounded corners, minimum size = 0.35cm] (4E) [above right = of 56] {};
                \node[circle, rounded corners, minimum size = 0.35cm] (60) [below left = of 51] {};
                \node[circle, rounded corners, minimum size = 0.35cm] (6E) [below right = of 56] {};
                \node[circle, rounded corners, minimum size = 0.35cm] (70) [below left = of 61] {};
                \node[circle, rounded corners, minimum size = 0.35cm] (7E) [below right = of 65] {};
                \node[circle, rounded corners, minimum size = 0.35cm] (80) [below left = of 71] {};
                \node[circle, rounded corners, minimum size = 0.35cm] (81) [below right = of 71] {};
                \node[circle, rounded corners, minimum size = 0.35cm] (82) [below right = of 72] {};
                \node[circle, rounded corners, minimum size = 0.35cm] (83) [below right = of 73] {};
                \node[circle, rounded corners, minimum size = 0.35cm] (8E) [below right = of 74] {};

                \path[every node/.style = {font = \sffamily\small}]
                (11) edge [right] node [left] {} (21)
                (11) edge [right] node [left] {} (22)
                (12) edge [right] node [left] {} (22)
                (12) edge [right] node [left] {} (23)
                (21) edge [right] node [left] {} (31)
                (21) edge [right] node [left] {} (32)
                (22) edge [right] node [left] {} (32)
                (22) edge [right] node [left] {} (33)
                (23) edge [right] node [left] {} (33)
                (23) edge [right] node [left] {} (34)
                (31) edge [right] node [left] {} (41)
                (31) edge [right] node [left] {} (42)
                (32) edge [right] node [left] {} (42)
                (32) edge [right] node [left] {} (43)
                (33) edge [right] node [left] {} (43)
                (33) edge [right] node [left] {} (44)
                (34) edge [right] node [left] {} (44)
                (34) edge [right] node [left] {} (45)
                (41) edge [right] node [left] {} (51)
                (41) edge [right] node [left] {} (52)
                (42) edge [right] node [left] {} (52)
                (42) edge [right] node [left] {} (53)
                (43) edge [right] node [left] {} (53)
                (43) edge [right] node [left] {} (54)
                (44) edge [right] node [left] {} (54)
                (44) edge [right] node [left] {} (55)
                (45) edge [right] node [left] {} (55)
                (45) edge [right] node [left] {} (56)
                (51) edge [right] node [left] {} (61)
                (52) edge [right] node [left] {} (61)
                (52) edge [right] node [left] {} (62)
                (53) edge [right] node [left] {} (62)
                (53) edge [right] node [left] {} (63)
                (54) edge [right] node [left] {} (63)
                (54) edge [right] node [left] {} (64)
                (55) edge [right] node [left] {} (64)
                (55) edge [right] node [left] {} (65)
                (56) edge [right] node [left] {} (65)
                (61) edge [right] node [left] {} (71)
                (62) edge [right] node [left] {} (71)
                (62) edge [right] node [left] {} (72)
                (63) edge [right] node [left] {} (72)
                (63) edge [right] node [left] {} (73)
                (64) edge [right] node [left] {} (73)
                (64) edge [right] node [left] {} (74)
                (65) edge [right] node [left] {} (74)
                (00) edge [right, dashed, gray] node [left] {} (11)
                (01) edge [right, dashed, gray] node [left] {} (11)
                (01) edge [right, dashed, gray] node [left] {} (12)
                (0E) edge [right, dashed, gray] node [left] {} (12)
                (10) edge [right, dashed, gray] node [left] {} (21)
                (1E) edge [right, dashed, gray] node [left] {} (23)
                (20) edge [right, dashed, gray] node [left] {} (31)
                (2E) edge [right, dashed, gray] node [left] {} (34)
                (30) edge [right, dashed, gray] node [left] {} (41)
                (3E) edge [right, dashed, gray] node [left] {} (45)
                (40) edge [right, dashed, gray] node [left] {} (51)
                (4E) edge [right, dashed, gray] node [left] {} (56)
                (51) edge [right, dashed, gray] node [left] {} (60)
                (56) edge [right, dashed, gray] node [left] {} (6E)
                (61) edge [right, dashed, gray] node [left] {} (70)
                (65) edge [right, dashed, gray] node [left] {} (7E)
                (71) edge [right, dashed, gray] node [left] {} (80)
                (71) edge [right, dashed, gray] node [left] {} (81)
                (72) edge [right, dashed, gray] node [left] {} (81)
                (72) edge [right, dashed, gray] node [left] {} (82)
                (73) edge [right, dashed, gray] node [left] {} (82)
                (73) edge [right, dashed, gray] node [left] {} (83)
                (74) edge [right, dashed, gray] node [left] {} (83)
                (74) edge [right, dashed, gray] node [left] {} (8E);
            \end{tikzpicture}
        }
        \caption{Motif for the model in Fig.~\ref{fig:triangle-grid}. Conditioning variables (shaded) form the Markov blanket of proposed variables (white). Dashed gray arrows are possible irrelevant dependencies.}
        \label{fig:triangle-grid-prop}
    \endminipage\hfill
    \minipage[b]{0.39\textwidth}
        \hspace{-0.85em}\includegraphics[scale = 0.26, trim = 5 5 0 0, clip]{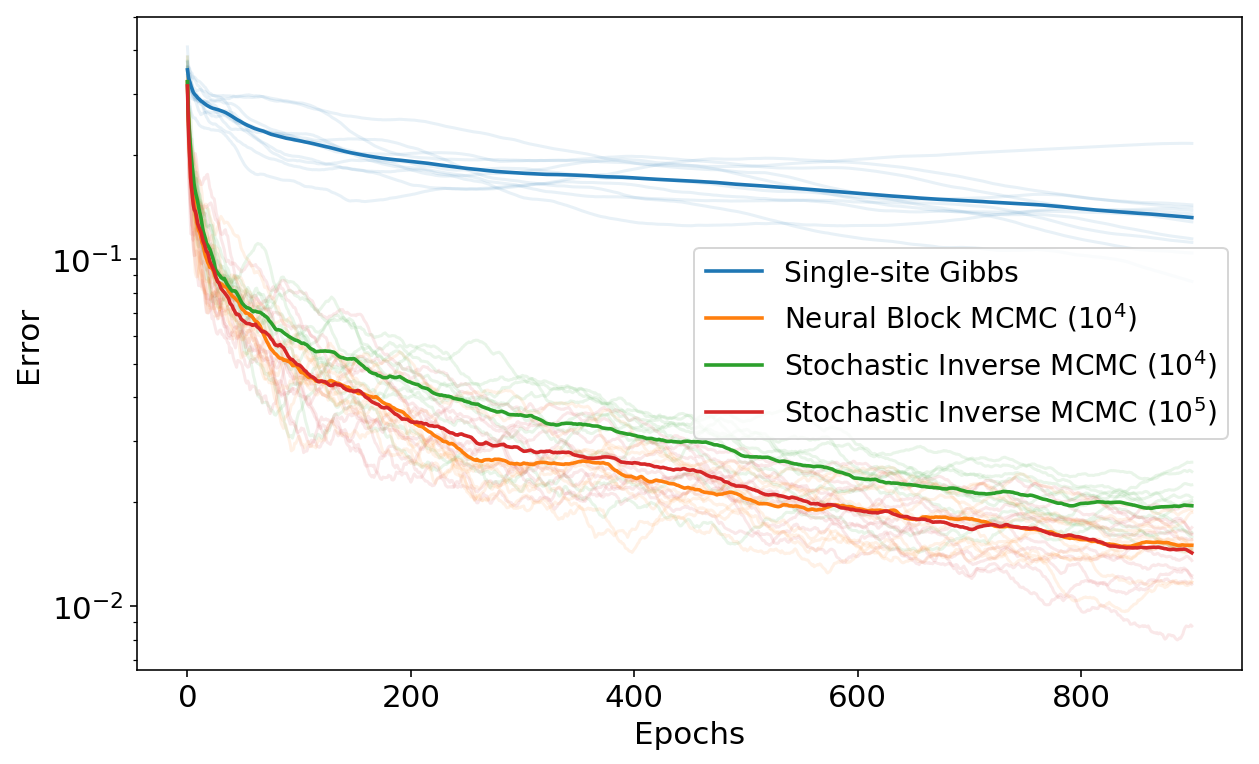}
        \caption{Error w.r.t.~epochs on the triangle model in Fig.~\ref{fig:triangle-grid}. Semitransparent lines show individual MCMC runs. Opaque lines show average error over $10$ MCMC runs for each algorithm. Numbers in parentheses are the amounts of training data. }
        \label{fig:grid_single_comp}
    \endminipage
\end{figure}%

\label{sec:stochastic_inverse_experiments}
Neural block proposals can also be used model-specifically by training only on instantiations within a particular model. In this subsection, we demonstrate that our method achieves comparable performance with a more complex task-specific MCMC method, Inverse MCMC~\citep{stuhlmuller2013learning}.

Figure~\ref{fig:triangle-grid} illustrates the triangle grid network used in this experiment, which is identical to what \citet{stuhlmuller2013learning} used to evaluate Inverse MCMC. For our method, we chose the motif shown in Fig.~\ref{fig:triangle-grid-prop}. \hide{The underlying MDN takes in assignments of conditioning variables and all relevant CPTs, then outputs a block proposal of $13$ variables. }The proposal is trained on all instantiations in this triangle model.

Inverse MCMC is an algorithm that builds auxiliary data structures offline to speed up inference. Given an inference task, it trains an inverse graph for each latent variable where the latent variable is at bottom and evidence variables are at top. These graphs are then used as MCMC proposals.

It is difficult to compare these two methods w.r.t.~time. While both methods require offline training, Inverse MCMC needs to train from scratch if the set of evidence nodes changes, yet neural block sampling only needs one-time training for different inference tasks on this model. 
In this experiment, for each inference epoch, both methods propose about $10.5$ values on average per latent variable. Figure~\ref{fig:grid_single_comp} shows a more meaningful comparison of error w.r.t.~epochs.\hide{ among single-site Gibbs, Neural Block Sampling, and Inverse MCMC with different amount of training data.} Our learned neural block proposal, trained using $10^4$ samples, achieves comparable performance with Inverse MCMC, which is trained using $10^5$ samples and builds model-specific data structures (inverse graphs).

\paragraph{Architecture}The underlying MDN has 161-1120-1120-224 network structure, mapping the CPTs of all $29$ motif variables and $16$ conditioning variable values to the proposal distribution of $13$ proposed variables as a mixture of $16$ components.
\paragraph{Inverse MCMC Setting} In this experiment, we run Inverse MCMC with frequency density estimator trained with posterior samples, proposal block size up to $20$ and Gibbs proposals precomputed, following the original approach of \citet{stuhlmuller2013learning}.

\end{document}